\documentclass{bmvc2k}

\usepackage{times}
\usepackage{soul}
\usepackage{url}
\usepackage{graphicx}
\usepackage{amsmath}
\usepackage{amsthm}
\usepackage{amsfonts} 
\usepackage{bbold}
\usepackage{booktabs}
\usepackage{algorithm}
\usepackage{algorithmic}
\usepackage{multirow}
\title{Proposal-based Temporal Action Localization with Point-level Supervision}

\addauthor{Yuan Yin}{yinyuan@iis.u-tokyo.ac.jp}{1}
\addauthor{Yifei Huang}{hyf@iis.u-tokyo.ac.jp}{1}
\addauthor{Ryosuke Furuta}{furuta@iis.u-tokyo.ac.jp}{1}
\addauthor{Yoichi Sato}{ysato@iis.u-tokyo.ac.jp}{1}

\addinstitution{
Institute of Industrial Science \\
The University of Tokyo \\
Tokyo, JP
}

\runninghead{Yin et al.}{Proposal-based Temporal Action Localization}

\begin{document}

\maketitle

\begin{abstract}
Point-level supervised temporal action localization (PTAL) aims at recognizing and localizing actions in untrimmed videos where only a single point (frame) within every action instance is annotated in training data. Without temporal annotations, most previous works adopt the multiple instance learning (MIL) framework, where the input video is segmented into non-overlapped short snippets, and action classification is performed independently on every short snippet. We argue that the MIL framework is suboptimal for PTAL because it operates on separated short snippets that contain limited temporal information. Therefore, the classifier only focuses on several easy-to-distinguish snippets instead of discovering the whole action instance without missing any relevant snippets. To alleviate this problem, we propose a novel method that localizes actions by generating and evaluating action proposals of flexible duration that involve more comprehensive temporal information. Moreover, we introduce an efficient clustering algorithm to efficiently generate dense pseudo labels that provide stronger supervision, and a fine-grained contrastive loss to further refine the quality of pseudo labels. Experiments show that our proposed method achieves competitive or superior performance to the state-of-the-art methods and some fully-supervised methods on four benchmarks: ActivityNet 1.3, THUMOS 14, GTEA, and BEOID datasets.
\let\thefootnote\relax\footnotetext{\hspace{-1.9em}Corresponding author: Yifei Huang}
\end{abstract}

\vspace{-5mm}
\section{Introduction}
\label{sec:intro}


Understanding what actions happened and when they happened in videos is beneficial to many applications, \textit{e.g.}, video surveillance~\cite{munro2020multi, li2023mseg3d}, human-computer interaction~\cite{masullo2022inertial, masullo2021no, chen2021integration,huang2020ego,yu2023fine}, and video analysis~\cite{wang2022tvnet, yang2022interact,huang2018predicting,huang2020improving,huang2022compound}. Also known as temporal action localization, most previous researches on this task focus on recognizing and localizing actions in videos under the fully-supervised setting~\cite{wang2021proposal, qing2021temporal, zhao2021video, zeng2021graph, xu2020g}, where frame-level carefully annotated videos are required. However, acquiring frame-wise annotations with precise start and end for all actions is labor-intensive. Meanwhile, the labels of precise start and end are potentially subjective, as it is hard to give a sensible definition for the action boundaries due to the ambiguity during the transition from one action to another~\cite{ma2020sf}.

To escalate practicability, many researchers have started to explore weaker levels of supervision, \textit{e.g.}, point-level supervision \cite{moltisanti2019action}.
In temporal action localization with point-level supervision, only one frame is annotated with its action category for each action instance. This annotated frame is also called the \textit{point-level annotation} for the action instance \cite{lee2021completeness}. Compared to frame-level supervision, point-level supervision significantly reduces labeling effort (from 300 seconds to about 50 seconds for a 1-minute video according to \cite{ma2020sf}). 

Most previous works \cite{ma2020sf,lee2021completeness} are built on top of the multiple instance learning (MIL) framework. In the MIL framework, the input video is segmented into short snippets (temporal segments), and the snippets that do not contain any point-level annotation are assigned as background. 
A classifier is trained on these snippets and then used to generate a class activation sequence (CAS) for the whole video. Then a model localizes the action by identifying the most activated parts of CAS via techniques such as thresholding.

However, methods with MIL-based framework essentially perform only snippet-level classification, which is not optimal for action localization from two perspectives: (1) the video snippets are usually short, and (2) the classifiers only treat video snippets independently. Therefore, the model will only focus on the easy-to-distinguish snippets instead of all the relevant snippets of an action instance. This does not correspond to the goal of predicting the accurate temporal boundaries that should cover the full duration of actions.  

To alleviate the downsides of the MIL framework, we design a proposal-based framework called Action Proposal Network (APN). 
Given an input video, APN first detects action boundaries by measuring the probabilities of each temporal location being the start or the end of an action instance. Then, APN generates action proposals via matching all pairs of possible starting and ending action boundaries. 
The generated action proposals have flexible lengths and contain more comprehensive temporal information compared to the short snippets \cite{ma2020sf,lee2021completeness}, which is beneficial to the classification and localization tasks. 
To classify and evaluate these action proposals simultaneously, APN learns a proposal-level representation using the features within and around proposals. Finally, inspired by the recent success in vision-language modeling, we classify and evaluate action proposals by predicting the text that describes the corresponding actions within proposals. We show that learning from this kind of additional natural language information can further help boosting the performance.

While the boundary detection scheme has been proven to be effective in the fully supervised action localization task~\cite{lin2019bmn}, since the point-level supervision is too weak for the APN to learn how to generate high-quality proposals, training APN poses a significant challenge. To overcome this issue, we propose a novel constrained k-medoids clustering algorithm to generate dense pseudo labels from the point-level annotations. We consistently update the pseudo labels during the training process, encouraging these pseudo labels to offer more precise information on action boundaries, thus providing guidance for the model to learn complete actions. Since the pseudo labels cannot accurately reflect action boundaries, we introduce a novel fine-grained contrastive loss that continuously refines the features. This loss is designed to fine-tune the action boundaries of pseudo labels and enhance the sensitivity of our proposed Action Proposal Network (APN) to action boundaries. The pseudo labels and contrastive loss work synergistically to provide more accurate and fine-tuned supervision to the model, which leads to better performance in terms of action localization.

\begin{figure*}[ht]
  \centering
   \includegraphics[width=\linewidth]{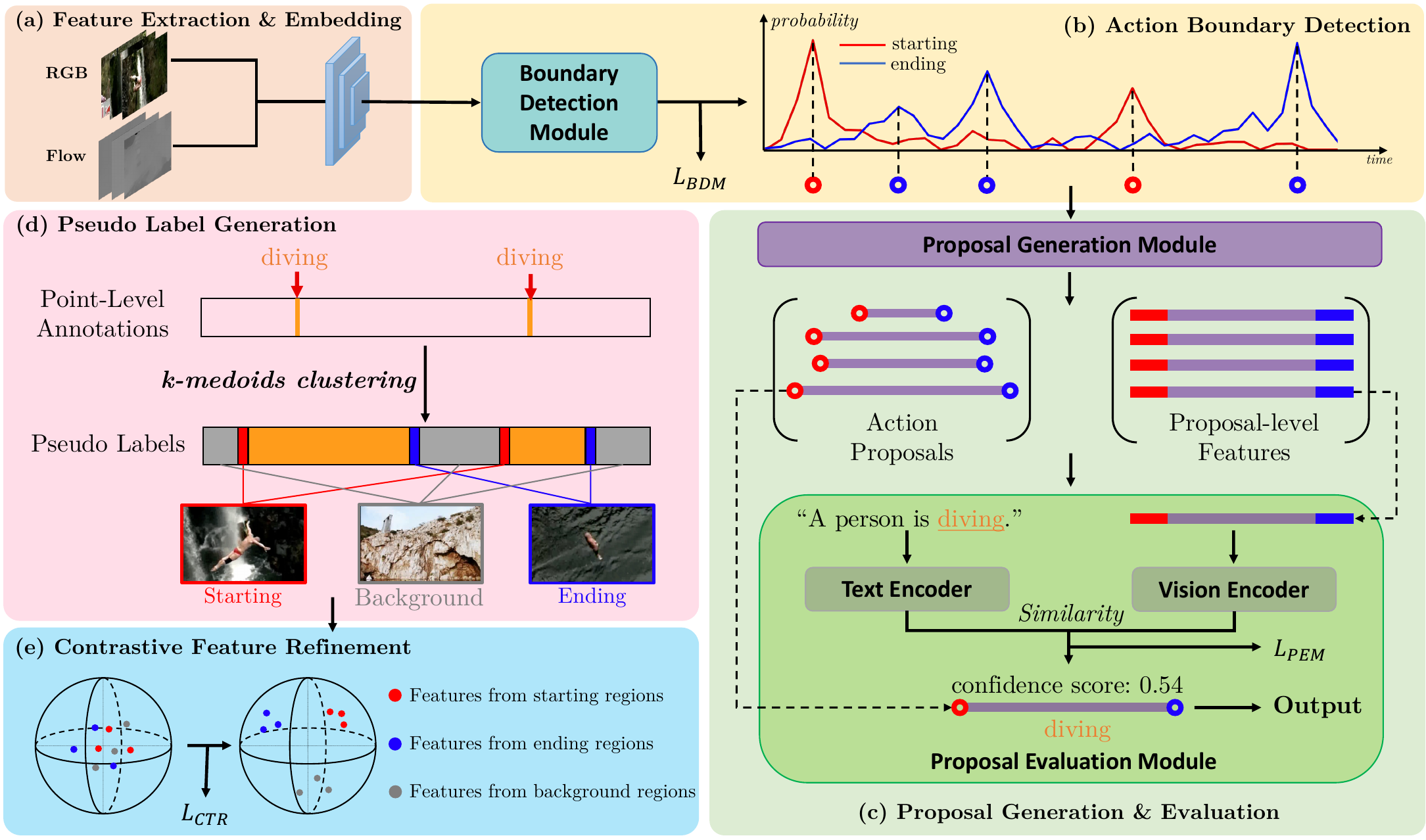}
   \vspace{-7mm}
   \caption{The overview of our method. (a) Given an input video, we extract its RGB and optical flow features and perform feature embedding. (b) On top of the embedded features, we measure the starting and ending probabilities of every temporal location. (c) Based on the possible starting and ending locations, we generate action proposals and then use proposal-level features to classify and evaluate the action proposals. After removing redundant proposals by post-processing, we output predictions. (d) To train the model, we propose a constrained k-medoids clustering algorithm to generate dense pseudo labels from point-level annotations and keep updating them. (e) We also introduce a contrastive loss to refine the embedded features from fine-grained regions.}
   \label{fig:overview}
\end{figure*}

The main contributions of our work are three-fold: (1) We propose a proposal-based method called Action Proposal Network (APN) for temporal action localization under point-level supervision. To the best of our knowledge, this is the first work that utilizes action proposals in PTAL. (2) We design an efficient clustering algorithm to generate dense pseudo labels in order to provide stronger supervision. We further introduce a novel contrastive loss to refine the boundaries of the generated pseudo labels and help APN detect action boundaries more accurately. (3) We demonstrate the effectiveness of our proposed method on four popular benchmarks, namely ActivityNet 1.3, THUMOS 14, GTEA, and BEOID. Experiments show that our method achieves competitive or superior performance to the state-of-the-art methods and even some fully-supervised counterparts. 

\vspace{-5mm}
\section{Related Work}

\textbf{Weakly-supervised Temporal Action Localization (WTAL)~} 
Instead of using expensive frame-level annotations, WTAL aims to train action localization models using weaker levels of supervision.
Early WTAL methods \cite{bojanowski2014weakly,huang2016connectionist,ding2018weakly} use transcript supervision, i.e., using ordered lists of actions in the videos as supervision. Since transcripts are only available in a few datasets, video-level supervision becomes the most common weak supervision in WTAL, which only indicates the presence of actions in a video. Based on the idea of multi-instance learning (MIL), \cite{liu2019completeness} used a multi-branch neural network and enforced each branch to discover diverse parts of actions so that the model can aggregate the information extracted from the various branches to make better predictions. \cite{junyu2022CVPR_FTCL} proposed a sequence-to-sequence comparing framework to explore the fine-grained distinction between action and background sequences, which further helps to separate the background from the action predictions. 

Recently, point-level supervision attracted extensive attention from researchers. Point-level supervision offers a single annotated frame for supervision. 
Many researchers find this type of supervision achieves a balance between labeling effort and the amount of information contained in this supervision. \cite{moltisanti2019action} designed a parameterized sampling function to sample relevant frames around the annotated frame and use them for training the action classifier. \cite{lee2021completeness} proposed to learn the completeness of the action from dense pseudo-labels by contrasting the action instances with surrounding background ones. 

The state-of-the-art works \cite{ma2020sf,lee2021completeness} are built on top of a similar MIL paradigm to the one used in many WTAL works \cite{zhang2021cola,shou2018autoloc,qu2021acm}, where the input video is segmented into short snippets (temporal segments), and each snippet is classified independently. This framework treats each snippet independently and ignores the temporal relationship between short snippets so that the resulting models will only focus on the most easy-to-discriminative snippet instead of all the relevant snippets. In contrast, our method generates flexible action proposals to locate actions and uses proposal-level features to classify the actions.

\noindent \textbf{Vision-language Modeling in Video Understanding~} With the recent success of vision-language modeling \cite{radford2021learning},  using natural language information to guide model training has been widely adopted in video understanding tasks. \cite{xu2021videoclip,xu2021vlm,yang2021taco,li2022bridge,wang2021actionclip} have shown that vision-language multi-modality can significantly improve the performance of FTAL models. We propose to leverage vision-language information in classifying and evaluating action proposals, which is rarely explored in previous WTAL approaches.

\vspace{-3mm}
\section{Method}

In this section, we first formulate the problem and then introduce the details of our method, including the pseudo label generation algorithm, the architecture of APN and the contrastive feature refinement. Finally, we discuss the training and inference process of our method. 
\vspace{-3mm}
\subsection{Problem Definition}

We formulate the task of PTAL as follows: First, we denote an input video $\mathcal{V}$ of $L$ frames as a sequence of frames $\mathcal{X} = \{ x_i\}_{i=1}^L$. Suppose the input video $\mathcal{V}$ contains $N_\mathcal{V}$ action instances, then the ground truth of these action instances can be denoted as: $\Phi_\mathcal{V} = \{\varphi_i = \left(t_i^s, t_i^e, c_i \right ) \}_{i=1}^{N_\mathcal{V}}$, where $t_i^s$, $t_i^e$ are the start and end time of $i$-th instance, respectively, $c_i$ are the corresponding action category. Our goal is to predict all the action instances in the input video. The resulting output should be: $\hat{\Phi}_\mathcal{V} = \{\hat{\varphi}_i = \left (\hat{t}_i^s, \hat{t}_i^e, \hat{c}_i, \hat{s}_i \right ) \}_{i=1}^{N_\mathcal{V}}$, where $\hat{t}^s$, $\hat{t}^e$, $\hat{c}$ are the corresponding estimated values, and $\hat{s}$ are the scores indicating how confident the model is about this prediction. During training, a set of point-level supervision is provided for every training video as $\Psi_V = \{\psi_i = \left (t_i^p, c_i\right ) \}_{i=1}^{N_\mathcal{V}}$, where $t_i^p$ is a single frame within the $i$-th action instance.
\vspace{-3mm}
\subsection{Feature Extraction and Embedding}

Following the previous works, we first segment the input video into 16-frame snippets, and then we use a pre-trained model, e.g. I3D \cite{carreira2017quo} to extract both the RGB features and optical flow features. We use a simple concatenation operation to fuse these two-stream features and get the video-level features $F \in \mathbb{R}^{T \times D}$, where $T$ is the number of segments, and $D$ is the feature dimensions. Since the extracted features $F$ are not designed for the TAL task, we use an embedding module consisting of a convolutional layer followed by ReLU activation to project $F$ into task-specific spaces, resulting in $X \in \mathbb{R}^{T \times D}$. 

\subsection{Pseudo label Generation} \label{pseudolabels}

 \begin{figure}[t]
  \centering
   \includegraphics[scale=0.4]{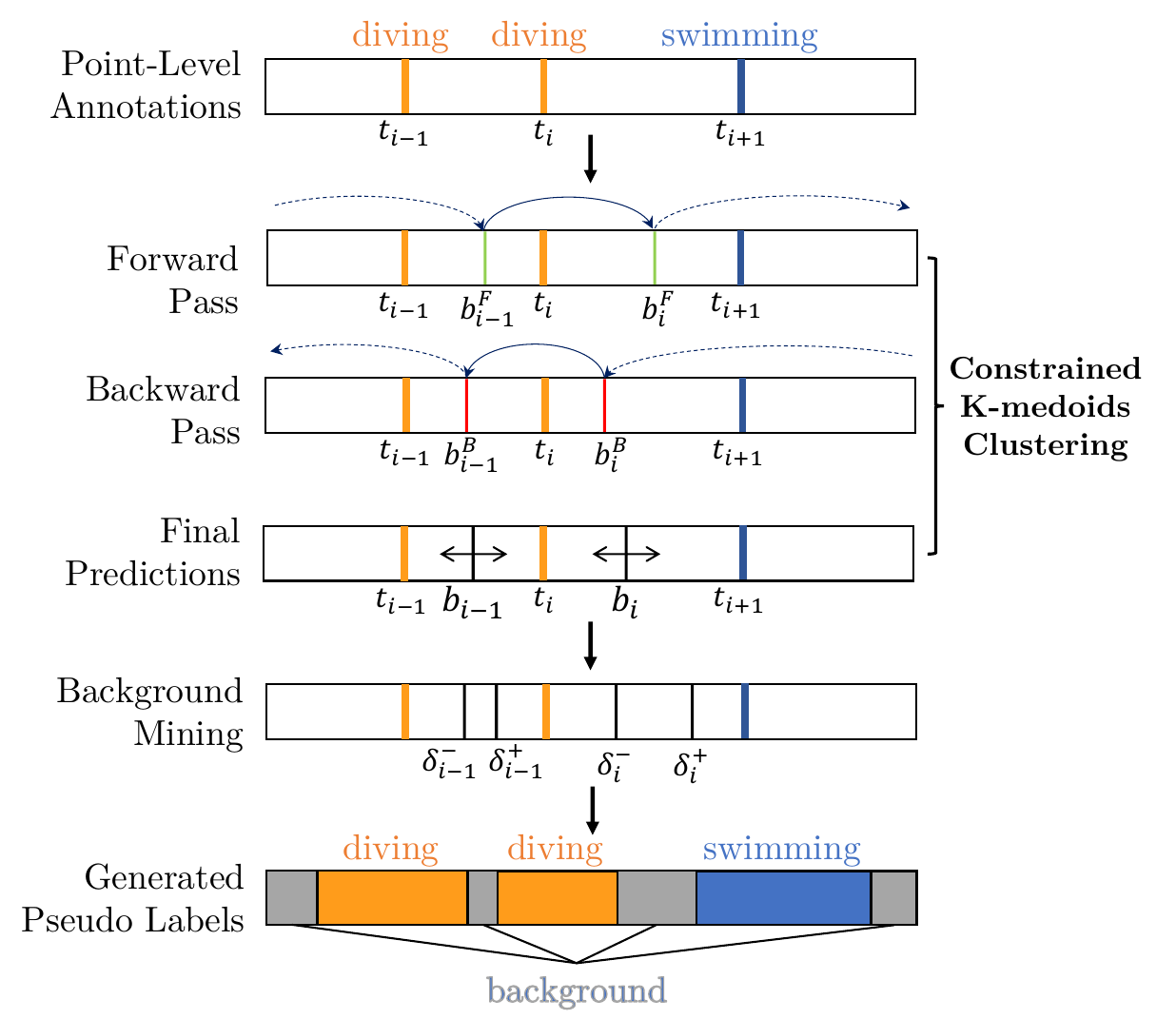}
   \caption{An overview of the pseudo label generation algorithm. Given point-level annotations, we propose a constrained k-medoids clustering algorithm that features forward-backward predictions to estimate the boundaries between actions. Then based on the predicted action boundaries, we mine background frames in order to provide more accurate pseudo labels. Finally, we output pseudo labeled action and background instances.}
   \label{fig:pseudo}
\end{figure}

Point-level annotations provide rather weak supervision because they do not contain action boundary information, which is crucial for training a TAL model. In order to make the supervision strong, we propose a novel algorithm to generate dense pseudo labels that can provide action boundary information. Although many previous works \cite{lee2021completeness,ma2020sf,huang2023weakly,huang2020mutual} also proposed various methods to generate pseudo labels, their methods largely depend on CAS which is not optimal for action localization tasks as discussed in Section~\ref{sec:intro}. 

Therefore, we propose a simple yet effective clustering algorithm to generate pseudo labels and update them during training. Given the observation that (1) the point-level annotations are inside action instances; (2) frames from the same action instance share similar visual features, it is natural to regard point-level annotations as cluster medoids and use a clustering algorithm to cluster surrounding frames. Then every cluster can represent an action instance in the video. However, in a conventional k-medoids algorithm, each point is assigned to the closest cluster medoid, this will cause the clusters not guaranteed to be temporally continuous, which does not fit the temporal action localization task. 

To ensure the frames' temporal continuity within each cluster, we instead find the boundary of each cluster. In other words, we find the frame that divides the frames between two consecutive point-level annotations into two clusters where the distance between frames and the cluster medoid is minimized. We show an overview of this pseudo label generation algorithm in Figure~\ref{fig:pseudo} and we further demonstrate the technical details of the proposed algorithm in the supplementary material. We run this clustering algorithm until convergence to get all the pseudo labeled action instances and background instances. During the training process, we update the pseudo labels after every $R$ iteration of the model training to provide better guidance to the model. 

\subsection{Action Proposal Network} \label{APN}

In order to localize actions with more comprehensive temporal information, we propose to replace the short snippets used in MIL-based methods with temporally longer action proposals. To this end, we design an Action Proposal Network (APN) that aims to generate action proposals of arbitrary temporal duration and with accurate action boundaries, and evaluate action proposals with reliable confidence scores. As shown in Figure~\ref{fig:overview}, APN is composed of three modules: \textit{Boundary Detection Module} (BDM), \textit{Proposal Generation Module} (PGM), and \textit{Proposal Evaluation Module} (PEM). 

\noindent \textbf{Boundary Detection Module~} 
BDM aims to evaluate the probability that each moment is the start or end point of an action from all $T$ moments of the input video. We adopt a multi-layer convolutional network with sigmoid activation that takes the embedded features $X \in \mathbb{R}^{T \times D}$ as input and outputs the starting probability sequence $\mathcal{P}^s = \{p_i^s\}_{i=1}^T$ and the ending probability sequence $\mathcal{P}^e = \{p_i^e\}_{i=1}^T$.

\noindent \textbf{Proposal Generation Module~} 
The goal of PGM is to generate action proposals according to the starting and ending probability sequences and sample proposal-level features. 
We follow the same method as described in \cite{lin2018bsn}, in which we traverse the starting and ending probability sequences to retrieve all possible starting boundaries $\mathcal{B}^s = \{\hat{t}_i^s\}_{i=1}^{N_s}$ and ending boundaries $\mathcal{B}^e = \{\hat{t}_i^e\}_{i=1}^{N_e}$. Then we iterate over $\mathcal{B}^s$ and $\mathcal{B}^e$ to find all the pairs $\left ( \hat{t}_i^s, \hat{t}_j^e \right )$ that $d = \hat{t}_i^s - \hat{t}_j^e \in [ d_{min}, d_{max} ]$, where $d_{min}$, $d_{max}$ are hyper-parameters that constraint the minimum and maximum temporal length of action proposals to avoid too long or too short proposals. These possible pairs are the resulting candidate action proposals, and we denote them as  $\hat{\Phi}_\mathcal{V} = \{\hat{\varphi}_i = \left (\hat{t}_i^s, \hat{t}_i^e \right ) \}_{i=1}^{\hat{N}_p}$, where $\hat{N}_p$ is the number of candidate proposals.

To sample proposal-level features for the $i$-th action proposal, we uniformly sample $N$ points from $[\hat{t}_i^s - d / 10, \hat{t}_i^e + d / 10]$, and use the features at these $N$ points $x_i \in \mathbb{R}^{N \times D}$ as the proposal-level features. These features contain rich visual and temporal information about proposals, and will be used to classify and evaluate the proposals.

\noindent \textbf{Proposal Evaluation Module~}
Most previous proposal-based TAL methods \cite{lin2020fast,yang2022temporal} follow a two-stage pipeline where the proposals are first evaluated, post-processed, and then classified. Instead, we integrate these two stages into one and show that these two tasks can mutually benefit from each other. The goal of PEM is to simultaneously (1) predict the action category and (2) evaluate the confidence score of each action proposal, to predict whether it contains a complete action instance. Inspired by the recent progress of vision-language modeling \cite{wang2021actionclip}, we regard this goal as a proposal-text matching problem. We first use prompt engineering \cite{li2022bridge,yang2023deco} to modify the textual action categories as descriptive texts and then compare the proposals with the prompted texts. Thus, if one action proposal can better match the text, there will be a higher probability that this proposal contains an action instance of the category described in the text. 

Technically, PEM consists of a vision encoder and a text encoder. We first turn all the action labels in the training data into descriptive texts using simple prompts. Given a textual action label $c$, a set of pre-defined prompts $\mathcal{Z}$, and a prompting function $f_{p}$, the prompted action label $c' = f_{p}(c, z)$, where $z \in \mathcal{Z}$. For example, if $c$ is \textit{``playing tennis"} and $z$ is \textit{``the man in the scene is \underline{\{   \}}"}, then the prompted text $c'$ will be \textit{``the man in the scene is \underline{\{playing tennis\}}"}. We use a text encoder to embed the prompted texts into textual action label representations $c_i^t$, and we learn vision representations using a vision encoder. The input to the vision encoder includes the proposal-based features $x_i \in \mathbb{R}^{N \times D}$ and an additional learnable token \texttt{[CLS]}. The encoded \texttt{[CLS]} token will pass through a simple multi-layer perceptron (MLP) model, which gives $\hat{c}_i \in \mathbb{R}^M$, to classify the action categories. The encoded vision representation $x_i^v$ will pass through the contrastive vision-language learning framework to infer the confidence scores.

Formally, for the $i$-th proposal, we use the cosine similarity between its representation $x_i^v$ and the $j$-th textual action label representation $c_j^t$ as the confidence score of the $i$-th proposal that contains the action instance of the $j$-th action category: $s_{ij} = s_{ij} = \frac{c_j^t\cdot x_i^v}{|c_j^t|\cdot|x_i^v|}$. In this way, PEM performs multi-task learning to simultaneously outputs classification results and confidence scores for each proposal: $\varphi = (\hat{t}_i^s, \hat{t}_i^e, \hat{c}_i, \hat{s}_i)$.

\subsection{Contrastive Feature Refinement} \label{contrast}

Since both pseudo label generation and BDM rely on the embedded features $X$, we introduce an additional fine-grained contrastive loss that refines the embedded features $X$. The motivation is that we want the features around the starting or ending moments of various instances belonging to the same action category to be close to each other, and we want to push them away from features of the background. For example, although the action \textit{``diving"} could occur in different contexts, it always starts with \textit{``jumping up from someplace"} and ends with \textit{``falling into the water"}. So we want the features of \textit{``jumping up from someplace"} or \textit{``falling into the water"} to be more similar to each other than to the features of the background. This helps the features around action boundaries to be more distinguishable than the features of background instances. 

To this end, we design a novel loss function based on InfoNCE \cite{oord2018representation} contrastive loss for contrastive feature refinement.
We first define the \textit{starting region} $r_g^s$ and \textit{ending region} $r_g^e$ of an action instance $\varphi = (t^s, t^e)$ as: $r_g^s = [t^s - d_g/10, t_s+d_g/10], r_g^e = [t^e - d_g/10, t_e+d_g/10]$, where $d_g = t^e - t^s$ is the length of the action instance.
To acquire the features of starting and ending regions, we average the embedded features within each starting region $r_g^s$ and ending region $r_g^e$ and denote them as $x_{ij}^s$ and $x_{ij}^e$ respectively, where $x_{ij}$ indicates the feature of the $i$-th instance from $j$-th action category. For the features of background regions, we average the features of each background instance in pseudo labels generated in Section~\ref{pseudolabels} and denote the features of the $k$-th background instance as $x^b_k$.

This loss is calculated when there are at least two action instances of the same action category in the input video in order to make the contrast. Given $x_{ij}^s, x_{ij}^e, x^b_k$, the contrastive loss $\mathcal{L}_{CTR}$ is then formulated as:
\begin{align}
    \mathcal{L}_{c}(x_{ij}, x^b) &= -\log\frac{\sum_{p\neq i} \exp(x_{ij}\cdot x_{pj}/\tau)}{\sum_{q\neq i} \exp(x_{ij}\cdot x_{qj}/\tau) + \sum_{k} \exp(x_{ij}\cdot x^b_k/\tau)} \\
    \mathcal{L}_{CTR} &= \sum_{j} \sum_{i} \left ( \mathcal{L}_{c}(x^s_{ij}, x^b) + \mathcal{L}_{c}(x^e_{ij}, x^b) \right )
\end{align}
where $\tau$ is the temperature parameter.

We name the proposed loss as ``fine-grained contrastive loss'' because it differs from the contrastive loss proposed in \cite{lee2021completeness}. The contrastive loss in \cite{lee2021completeness} is designed to let the model learn action completeness by contrasting action instances with background ones. It encourages the features of \textbf{entire} action instances from the same action class to be closer than the features of other background instances. In our model, the Boundary Detection Module (BDM) predicts the starting and ending points of action proposals, which largely determine the completeness of predicted actions. Therefore, instead of operating on the level of the entire action instances, we choose a more \textbf{fine-grained} level: to contrast between the features around starting/ending points and the background ones. In this way, we finetune the input features to BDM so that we can get more accurate predictions of action boundaries.

\subsection{Training}

For the training of BDM, we use the same training objectives $\mathcal{L}_{BDM}$ as described in \cite{lin2018bsn}. And for PEM, we first discard all proposals with no or more than one point-level annotation that falls within them. In other words, we ensure that all the proposals involved in training contain exactly one point-level annotation. We use the sum of binary cross-entropy loss between every proposal's classification results and confidence scores $(\hat{c}_i, \hat{s}_i)$ and their labels $(c_i, s_i)$ as the training objective for PEM: 
\begin{equation}
	\mathcal{L}_{PEM} = -\frac{1}{N_p}\sum_{i=1}^{N_p} \left( c_i \cdot log( \hat{c}_i) + s_i \cdot log( \hat{s}_i) \right )
\end{equation}

Finally, we train APN end-to-end with all the loss functions from above:
\begin{equation}
	\mathcal{L}_{total} = \mathcal{L}_{BDM} + \lambda_1 \cdot \mathcal{L}_{PEM} + \lambda_2 \cdot \mathcal{L}_{CTR}
\end{equation}
where weights $\lambda_1$ and $\lambda_2$ are used to balance the influence of different loss functions.

\section{Experiments}

\subsection{Datasets and Setup}

 \textbf{Datasets~} We evaluate our method on four popular datasets, namely ActivityNet 1.3 \cite{caba2015activitynet}, THUMOS 14 \cite{THUMOS14}, GTEA \cite{lei2018temporal} and BEOID \cite{damen2014you}. The ActivityNet 1.3 is a large video understanding dataset that contains 10,024 training videos, 4,926 validation videos, and 5,044 test videos from 200 action categories. The durations of action instances distribute evenly from short to long, so it is well-suited to verify the effectiveness of our method. The THUMOS 14 contains 200 validation videos and 213 test videos from 20 action categories. The GTEA dataset contains 28 videos from 7 action classes. The BEOID dataset contains 58 videos from 34 action classes. Since all these datasets do not have official point-level annotations, we follow \cite{moltisanti2019action} to automatically generate point-level annotations and use them for training.

 \noindent \textbf{Evaluation Metrics~} To evaluate our method, we adopt mean average precision (mAP) under various thresholds, which is commonly used in TAL. For a fair comparison, we use the same setting as previous works: For experiments on ActivityNet 1.3, we use mAP calculated with tIoU thresholds between 0.5 and 0.95 with the step size of 0.05; for experiments on THUMOS 14, GTEA and BEOID, we use mAP computed with tIoU thresholds between 0.1 and 0.7 with the step size of 0.1.

\noindent \textbf{Implementation~} As for implementation, We use the two-stream I3D \cite{carreira2017quo} network pre-trained on Kinetics-400 dataset \cite{carreira2017quo} as the feature extractor. We adopt the TV-L1 algorithm \cite{wedel2009improved} to extract optical flow from input videos. The output features are 1024-dim vectors for each modality. We update pseudo labels every $R=10$ iterations. We optimize our model with the Adam \cite{kingma2014adam} optimizer and a learning rate of 0.0001. Hyper-parameters are set as: $N=32, \tau=0.1, \lambda_1=1, \lambda_2=0.1$. On both datasets, we train our model with a mini-batch size of 16 for 30 epochs. The threshold for NMS is set to 0.5.

\subsection{Comparison with State-of-the-art Methods} \label{compsota}

\begin{table*}[ht]
  \centering
  \caption{Comparison with state-of-the-art TAL methods under different levels of supervision on ActivityNet 1.3 dataset. The AVG column shows the averaged mAP under the thresholds [0.5:0.05:0.95].}
  \scalebox{0.7}{
    \begin{tabular}{c|l|ccc|cr}
\cmidrule{1-6}    \multirow{2}[2]{*}{Supervision} & \multicolumn{1}{c|}{\multirow{2}[2]{*}{Method}} & \multicolumn{3}{c|}{mAP@IoU (\%)} & \multirow{2}[2]{*}{AVG} &  \\
          &       & 0.5   & 0.75  & 0.95  &       &  \\
\cmidrule{1-6}    \multirow{5}[2]{*}{Frame-level (FTAL)} & BMN \cite{lin2019bmn} & 50.1  & 34.8  & 8.3   & 33.9  &  \\
          & BSN \cite{lin2018bsn} & 46.5  & 30.0  & 8.0   & 30.0  &  \\
          & G-TAD \cite{xu2020g} & 50.4  & 34.6  & 9.0   & 34.1  &  \\
          & TAGS \cite{nag2022temporal} & 56.3  & 36.8  & 9.6   & 36.5  &  \\
\cmidrule{1-6}    \multirow{5}[2]{*}{Video-level (WTAL)} & FAC-Net \cite{huang2021foreground} & 37.6  & 24.2  & 6.0   & 24.0  &  \\
          & ACM-Net \cite{qu2021acm} & 37.6  & 24.7  & 6.5   & 24.4  &  \\
          & FTCL \cite{junyu2022CVPR_FTCL} & 40.0  & 24.3  & 6.4   & 24.8  &  \\
          & ASM-Loc \cite{he2022asm} & 41.0  & 24.9  & 6.2   & 25.1  &  \\
\cmidrule{1-6}    \multirow{2}[2]{*}{Point-level (PTAL)} & LACP \cite{lee2021completeness} & 40.4  & 24.6  & 5.7   & 25.1  &  \\
          & Ours & \textbf{48.3} & \textbf{27.8} & \textbf{7.0} & \textbf{29.1} &  \\
\cmidrule{1-6}    \end{tabular}%
    }
  \label{tab:activitynet}%
  \vspace{-0.3cm}
\end{table*}%

We compare our method with the state-of-the-art TAL methods under different types of supervision on the ActivityNet 1.3 dataset, and the results are shown in Table~\ref{tab:activitynet}. As shown in Table~\ref{tab:activitynet}, our proposed method achieves significantly better performance under all the IoU thresholds than the previous state-of-the-art method on PTAL, which shows the effectiveness of our method. Also, our method outperforms video-level WTAL methods by a large margin while maintaining a comparable labeling effort, which indicates that point-level supervision brings more information and provides more positive guidance to the model. It is worth noting that our method is comparable with several FTAL methods. 

We also conduct experiments on the THUMOS 14, GTEA, and BEOID datasets, and the results are shown in Table~\ref{tab:THUMOS14} and Table~\ref{tab:GTEA} . Our method performs significantly better on GTEA and BEOID datasets while performing comparably with the state-of-the-art methods on THUMOS 14 dataset. 

\begin{table}[htbp]
  \centering
  \caption{Comparison with state-of-the-art PTAL methods on THUMOS 14 dataset.}
  \scalebox{0.8}{
\begin{tabular}{c|ccccccc|cc}
\hline
\multirow{2}{*}{Method}        & \multicolumn{7}{c|}{mAP@IoU (\%)}                                                                             & \multirow{2}{*}{\begin{tabular}[c]{@{}c@{}}AVG\\ (0.1:0.5)\end{tabular}} & \multirow{2}{*}{\begin{tabular}[c]{@{}c@{}}AVG\\ (0.3:0.7)\end{tabular}} \\
                               & 0.1           & 0.2           & 0.3           & 0.4           & 0.5           & 0.6           & 0.7           &                                                                              &                                                                              \\ \hline
SF-Net\cite{ma2020sf}          & 68.3          & 62.3          & 52.8          & 42.2          & 30.5          & 20.6          & 12.0          & 51.2                                                                         & 31.6                                                                         \\
DCST\cite{ju2021divide}        & 72.3          & 64.7          & 58.2          & 47.1          & 35.9          & 23.0          & 12.8          & 55.6                                                                         & 35.4                                                                         \\
LACP\cite{lee2021completeness} & 75.7          & 71.4          & 64.6          & \textbf{56.5} & \textbf{45.3} & \textbf{34.5} & \textbf{21.8} & 62.7                                                                         & \textbf{44.5}                                                                \\
Ours                           & \textbf{77.1} & \textbf{72.6} & \textbf{65.9} & 54.4          & 44.9          & 33.1          & 20.2          & \textbf{63.3}                                                                & 43.9                                                                         \\ \hline
\end{tabular}
    }
  \label{tab:THUMOS14}%
\end{table}%

\begin{table}[htbp]
  \centering
  \caption{Comparison with state-of-the-art PTAL methods on GTEA and BEOID dataset. The AVG column shows the averaged mAP under the thresholds [0.1:0.1:0.7].}
  \scalebox{0.8}{
\begin{tabular}{c|ccccc|ccccc}
\hline
\multirow{3}{*}{Method}        & \multicolumn{5}{c|}{GTEA}                                                                                 & \multicolumn{5}{c}{BEOID}                                                                                 \\ \cline{2-11} 
                               & \multicolumn{4}{c|}{mAP@IoU (\%)}                                                  & \multirow{2}{*}{AVG} & \multicolumn{4}{c|}{mAP@IoU (\%)}                                                  & \multirow{2}{*}{AVG} \\
                               & 0.1           & 0.3           & 0.5           & \multicolumn{1}{c|}{0.7}           &                      & 0.1           & 0.3           & 0.5           & \multicolumn{1}{c|}{0.7}           &                      \\ \hline
SF-Net\cite{ma2020sf}          & 58.0          & 37.9          & 19.3          & \multicolumn{1}{c|}{11.9}          & 31.0                 & 62.9          & 40.6          & 16.7          & \multicolumn{1}{c|}{3.5}           & 30.1                 \\
DCST\cite{ju2021divide}        & 59.7          & 38.3          & 21.9          & \multicolumn{1}{c|}{18.1}          & 33.7                 & 63.2          & 46.8          & 20.9          & \multicolumn{1}{c|}{5.8}           & 34.9                 \\
LACP\cite{lee2021completeness} & \textbf{63.9} & \textbf{55.7} & 33.9          & \multicolumn{1}{c|}{20.8}          & 43.5                 & 76.9          & 61.4          & 42.7          & \multicolumn{1}{c|}{25.1}          & 51.8                 \\
Ours                           & 63.1          & 52.1          & \textbf{37.3} & \multicolumn{1}{c|}{\textbf{22.2}} & \textbf{45.1}        & \textbf{78.2} & \textbf{65.3} & \textbf{45.1} & \multicolumn{1}{c|}{\textbf{26.6}} & \textbf{54.2}        \\ \hline
\end{tabular}
    }
  \label{tab:GTEA}%
\end{table}%

\subsection{Analysis}

\textbf{Pseudo Label Generation~} In Table~\ref{tab:pseudolabel}, we evaluate the quality of generated pseudo labels. Specifically, we compare the mAP at threshold $0.5$ of the pseudo labels generated in different iterations of training. We can see that the mAPs of pseudo labels improve continuously as the training process goes on, regardless of the potential impact of $\mathcal{L}_{CTR}$. This verifies the effectiveness of our proposed pseudo label generation algorithm and the refinement brought by the continuous updating during training. 

\begin{table}[]
\centering
\parbox{.45\linewidth}{
  \centering
  \caption{Influence of progressive updating and the fine-grained contrastive loss $\mathcal{L}_{CTR}$ on the quality of pseudo labels on ActivityNet 1.3.}
  \scalebox{0.7}{
    \begin{tabular}{cccccr}
\cmidrule{1-5}    Training Epochs & 0     & 10    & 20    & 30    &  \\
\cmidrule{1-5} 
    mAP@0.5 (w/o $\mathcal{L}_{CTR}$) & 33.4  & 38.9  & 41.3  & 41.6  &  \\
     mAP@0.5 (w $\mathcal{L}_{CTR}$) & 33.4  & \textbf{42.3}  & \textbf{45.1}  & \textbf{45.8}  &  \\
\cmidrule{1-5}    \end{tabular}%
    }
  \label{tab:pseudolabel}%
}
\hfill
\parbox{.45\linewidth}{
  \centering
  \caption{Influence of the fine-grained contrastive loss $\mathcal{L}_{CTR}$ on the model's performance on ActivityNet 1.3.}
  \scalebox{0.7}{
    \begin{tabular}{ccccc}
    \toprule
    \multicolumn{1}{c}{\multirow{2}[2]{*}{$\mathcal{L}_{CTR}$}} & \multicolumn{3}{c}{mAP@IoU (\%)} & \multirow{2}[2]{*}{AVG} \\
    \multicolumn{1}{c}{} & 0.5   & 0.75  & \multicolumn{1}{c}{0.95} &  \\
    \midrule
    w/o   & \multicolumn{1}{r}{44.1} & \multicolumn{1}{r}{23.9} & \multicolumn{1}{r}{5.9} & \multicolumn{1}{r}{26.3} \\
    w     & \textbf{48.3} & \textbf{27.8} & \textbf{7.0} & \textbf{29.1} \\
    \bottomrule
    \end{tabular}%
    }
  \label{tab:ctrloss}%
}
\end{table}%

\noindent \textbf{Effectiveness of Contrastive Feature Refinement~} To verify the effectiveness of contrastive feature refinement, we show the influence of $\mathcal{L}_{CTR}$ on the quality of pseudo labels in Table~\ref{tab:pseudolabel} and the model's performance in Table~\ref{tab:ctrloss}. When trained with $\mathcal{L}_{CTR}$, the generated pseudo labels are significantly more precise than the ones trained without $\mathcal{L}_{CTR}$, indicating that $\mathcal{L}_{CTR}$ is helpful in improving the precision of pseudo label. In addition, $\mathcal{L}_{CTR}$ consistently improves the performance of our method at different IoU thresholds. These all prove that the contrastive feature refinement brought by $\mathcal{L}_{CTR}$ is beneficial to achieving more accurate temporal action localization.

\noindent \textbf{Vision-language Modeling~} 
We conduct an ablation study to validate the effectiveness of our vision-language modeling. Due to the page limit, we place the detailed experiment analysis in our supplementary material.

\noindent \textbf{Limitations and Future Work~}
We also analyze the limitations and promising directions of improvement. We put the analysis in our supplementary material due to the space limit.

\vspace{-5mm}
\section{Conclusion}

In this paper, we propose a novel proposal-based framework for PTAL, which can flexibly generate action proposals with more temporal information compared to previous MIL-based methods. We propose an efficient pseudo label generation algorithm to provide action boundary information to our model, which narrows the performance gap between weakly-supervised methods and fully-supervised methods. We further introduce a fine-grained contrastive loss to refine the features, improving the quality of pseudo labels and the accuracy of action localization. Experiments on four benchmarks: ActivityNet 1.3, THUMOS 14, GTEA, and BEOID prove that the proposed pseudo labels generation algorithm and contrastive loss help our model to localize action instances better. Our method achieves impressive results on all four benchmarks.

\small{
\noindent{\textbf{Acknowledgement}}~ This work is supported by JSPS KAKENHI Grant Number JP20H04205 and JP22KF0119.}
\bibliography{egbib}

\begin{thebibliography}{53}
\providecommand{\natexlab}[1]{#1}
\providecommand{\url}[1]{\texttt{#1}}
\expandafter\ifx\csname urlstyle\endcsname\relax
  \providecommand{\doi}[1]{doi: #1}\else
  \providecommand{\doi}{doi: \begingroup \urlstyle{rm}\Url}\fi

\bibitem[Bojanowski et~al.(2014)Bojanowski, Lajugie, Bach, Laptev, Ponce,
  Schmid, and Sivic]{bojanowski2014weakly}
Piotr Bojanowski, R{\'e}mi Lajugie, Francis Bach, Ivan Laptev, Jean Ponce,
  Cordelia Schmid, and Josef Sivic.
\newblock Weakly supervised action labeling in videos under ordering
  constraints.
\newblock In \emph{European Conference on Computer Vision}, pages 628--643.
  Springer, 2014.

\bibitem[Caba~Heilbron et~al.(2015)Caba~Heilbron, Escorcia, Ghanem, and
  Carlos~Niebles]{caba2015activitynet}
Fabian Caba~Heilbron, Victor Escorcia, Bernard Ghanem, and Juan Carlos~Niebles.
\newblock Activitynet: A large-scale video benchmark for human activity
  understanding.
\newblock In \emph{Proceedings of the ieee conference on computer vision and
  pattern recognition}, pages 961--970, 2015.

\bibitem[Carreira and Zisserman(2017)]{carreira2017quo}
Joao Carreira and Andrew Zisserman.
\newblock Quo vadis, action recognition? a new model and the kinetics dataset.
\newblock In \emph{proceedings of the IEEE Conference on Computer Vision and
  Pattern Recognition}, pages 6299--6308, 2017.

\bibitem[Chen et~al.(2021)Chen, Nakamura, Kondo, Damen, and
  Mayol-Cuevas]{chen2021integration}
Longfei Chen, Yuichi Nakamura, Kazuaki Kondo, Dima Damen, and Walterio
  Mayol-Cuevas.
\newblock Integration of experts' and beginners' machine operation experiences
  to obtain a detailed task model.
\newblock \emph{IEICE TRANSACTIONS on Information and Systems}, 104\penalty0
  (1):\penalty0 152--161, 2021.

\bibitem[Damen et~al.(2014)Damen, Leelasawassuk, Haines, Calway, and
  Mayol-Cuevas]{damen2014you}
Dima Damen, Teesid Leelasawassuk, Osian Haines, Andrew Calway, and Walterio~W
  Mayol-Cuevas.
\newblock You-do, i-learn: Discovering task relevant objects and their modes of
  interaction from multi-user egocentric video.
\newblock In \emph{BMVC}, volume~2, page~3, 2014.

\bibitem[Ding and Xu(2018)]{ding2018weakly}
Li~Ding and Chenliang Xu.
\newblock Weakly-supervised action segmentation with iterative soft boundary
  assignment.
\newblock In \emph{Proceedings of the IEEE Conference on Computer Vision and
  Pattern Recognition}, pages 6508--6516, 2018.

\bibitem[Gao et~al.(2022)Gao, Chen, and Xu]{junyu2022CVPR_FTCL}
Junyu Gao, Mengyuan Chen, and Changsheng Xu.
\newblock Fine-grained temporal contrastive learning for weakly-supervised
  temporal action localization.
\newblock In \emph{IEEE/CVF Conference on Computer Vision and Pattern
  Recognition (CVPR)}, 2022.

\bibitem[He et~al.(2022)He, Yang, Kang, Cheng, Zhou, and
  Shrivastava]{he2022asm}
Bo~He, Xitong Yang, Le~Kang, Zhiyu Cheng, Xin Zhou, and Abhinav Shrivastava.
\newblock Asm-loc: Action-aware segment modeling for weakly-supervised temporal
  action localization.
\newblock In \emph{Proceedings of the IEEE/CVF Conference on Computer Vision
  and Pattern Recognition}, pages 13925--13935, 2022.

\bibitem[Huang et~al.(2016)Huang, Fei-Fei, and Niebles]{huang2016connectionist}
De-An Huang, Li~Fei-Fei, and Juan~Carlos Niebles.
\newblock Connectionist temporal modeling for weakly supervised action
  labeling.
\newblock In \emph{European Conference on Computer Vision}, pages 137--153,
  2016.

\bibitem[Huang et~al.(2021)Huang, Wang, and Li]{huang2021foreground}
Linjiang Huang, Liang Wang, and Hongsheng Li.
\newblock Foreground-action consistency network for weakly supervised temporal
  action localization.
\newblock In \emph{Proceedings of the IEEE/CVF International Conference on
  Computer Vision}, pages 8002--8011, 2021.

\bibitem[Huang et~al.(2018)Huang, Cai, Li, and Sato]{huang2018predicting}
Yifei Huang, Minjie Cai, Zhenqiang Li, and Yoichi Sato.
\newblock Predicting gaze in egocentric video by learning task-dependent
  attention transition.
\newblock In \emph{European Conference on Computer Vision}, pages 754--769,
  2018.

\bibitem[Huang et~al.(2020{\natexlab{a}})Huang, Cai, Li, Lu, and
  Sato]{huang2020mutual}
Yifei Huang, Minjie Cai, Zhenqiang Li, Feng Lu, and Yoichi Sato.
\newblock Mutual context network for jointly estimating egocentric gaze and
  action.
\newblock \emph{IEEE Transactions on Image Processing}, 29:\penalty0
  7795--7806, 2020{\natexlab{a}}.

\bibitem[Huang et~al.(2020{\natexlab{b}})Huang, Cai, and Sato]{huang2020ego}
Yifei Huang, Minjie Cai, and Yoichi Sato.
\newblock An ego-vision system for discovering human joint attention.
\newblock \emph{IEEE Transactions on Human-Machine Systems}, 50\penalty0
  (4):\penalty0 306--316, 2020{\natexlab{b}}.

\bibitem[Huang et~al.(2020{\natexlab{c}})Huang, Sugano, and
  Sato]{huang2020improving}
Yifei Huang, Yusuke Sugano, and Yoichi Sato.
\newblock Improving action segmentation via graph-based temporal reasoning.
\newblock In \emph{Proceedings of the IEEE/CVF conference on computer vision
  and pattern recognition}, pages 14024--14034, 2020{\natexlab{c}}.

\bibitem[Huang et~al.(2022)Huang, Yang, and Sato]{huang2022compound}
Yifei Huang, Lijin Yang, and Yoichi Sato.
\newblock Compound prototype matching for few-shot action recognition.
\newblock In \emph{European Conference on Computer Vision}, pages 351--368,
  2022.

\bibitem[Huang et~al.(2023)Huang, Yang, and Sato]{huang2023weakly}
Yifei Huang, Lijin Yang, and Yoichi Sato.
\newblock Weakly supervised temporal sentence grounding with uncertainty-guided
  self-training.
\newblock In \emph{Proceedings of the IEEE/CVF Conference on Computer Vision
  and Pattern Recognition}, pages 18908--18918, 2023.

\bibitem[Jiang et~al.(2014)Jiang, Liu, Roshan~Zamir, Toderici, Laptev, Shah,
  and Sukthankar]{THUMOS14}
Y.-G. Jiang, J.~Liu, A.~Roshan~Zamir, G.~Toderici, I.~Laptev, M.~Shah, and
  R.~Sukthankar.
\newblock {THUMOS} challenge: Action recognition with a large number of
  classes.
\newblock \url{http://crcv.ucf.edu/THUMOS14/}, 2014.

\bibitem[Ju et~al.(2021)Ju, Zhao, Chen, Zhang, Wang, and Tian]{ju2021divide}
Chen Ju, Peisen Zhao, Siheng Chen, Ya~Zhang, Yanfeng Wang, and Qi~Tian.
\newblock Divide and conquer for single-frame temporal action localization.
\newblock In \emph{Proceedings of the IEEE/CVF International Conference on
  Computer Vision}, pages 13455--13464, 2021.

\bibitem[Kingma and Ba(2014)]{kingma2014adam}
Diederik~P Kingma and Jimmy Ba.
\newblock Adam: A method for stochastic optimization.
\newblock \emph{arXiv preprint arXiv:1412.6980}, 2014.

\bibitem[Lee and Byun(2021)]{lee2021completeness}
Pilhyeon Lee and Hyeran Byun.
\newblock Learning action completeness from points for weakly-supervised
  temporal action localization.
\newblock In \emph{Proceedings of the IEEE/CVF International Conference on
  Computer Vision}, pages 13648--13657, 2021.

\bibitem[Lei and Todorovic(2018)]{lei2018temporal}
Peng Lei and Sinisa Todorovic.
\newblock Temporal deformable residual networks for action segmentation in
  videos.
\newblock In \emph{Proceedings of the IEEE conference on computer vision and
  pattern recognition}, pages 6742--6751, 2018.

\bibitem[Li et~al.(2023)Li, Dai, Han, and Ding]{li2023mseg3d}
Jiale Li, Hang Dai, Hao Han, and Yong Ding.
\newblock Mseg3d: Multi-modal 3d semantic segmentation for autonomous driving.
\newblock \emph{Proceedings of the IEEE/CVF conference on computer vision and
  pattern recognition}, 2023.

\bibitem[Li et~al.(2022)Li, Chen, Duan, Hu, Feng, Zhou, and Lu]{li2022bridge}
Muheng Li, Lei Chen, Yueqi Duan, Zhilan Hu, Jianjiang Feng, Jie Zhou, and Jiwen
  Lu.
\newblock Bridge-prompt: Towards ordinal action understanding in instructional
  videos.
\newblock In \emph{Proceedings of the IEEE/CVF Conference on Computer Vision
  and Pattern Recognition}, pages 19880--19889, 2022.

\bibitem[Lin et~al.(2020)Lin, Li, Wang, Tai, Luo, Cui, Wang, Li, Huang, and
  Ji]{lin2020fast}
Chuming Lin, Jian Li, Yabiao Wang, Ying Tai, Donghao Luo, Zhipeng Cui, Chengjie
  Wang, Jilin Li, Feiyue Huang, and Rongrong Ji.
\newblock Fast learning of temporal action proposal via dense boundary
  generator.
\newblock In \emph{Proceedings of the AAAI Conference on Artificial
  Intelligence}, volume~34, pages 11499--11506, 2020.

\bibitem[Lin et~al.(2018)Lin, Zhao, Su, Wang, and Yang]{lin2018bsn}
Tianwei Lin, Xu~Zhao, Haisheng Su, Chongjing Wang, and Ming Yang.
\newblock Bsn: Boundary sensitive network for temporal action proposal
  generation.
\newblock In \emph{Proceedings of the European conference on computer vision
  (ECCV)}, pages 3--19, 2018.

\bibitem[Lin et~al.(2019)Lin, Liu, Li, Ding, and Wen]{lin2019bmn}
Tianwei Lin, Xiao Liu, Xin Li, Errui Ding, and Shilei Wen.
\newblock Bmn: Boundary-matching network for temporal action proposal
  generation.
\newblock In \emph{Proceedings of the IEEE/CVF international conference on
  computer vision}, pages 3889--3898, 2019.

\bibitem[Liu et~al.(2019)Liu, Jiang, and Wang]{liu2019completeness}
Daochang Liu, Tingting Jiang, and Yizhou Wang.
\newblock Completeness modeling and context separation for weakly supervised
  temporal action localization.
\newblock In \emph{Proceedings of the IEEE/CVF Conference on Computer Vision
  and Pattern Recognition}, pages 1298--1307, 2019.

\bibitem[Ma et~al.(2020)Ma, Zhu, Yang, Zha, Kundu, Feiszli, and Shou]{ma2020sf}
Fan Ma, Linchao Zhu, Yi~Yang, Shengxin Zha, Gourab Kundu, Matt Feiszli, and
  Zheng Shou.
\newblock Sf-net: Single-frame supervision for temporal action localization.
\newblock In \emph{European conference on computer vision}, pages 420--437.
  Springer, 2020.

\bibitem[Masullo et~al.(2021)Masullo, Perrett, Damen, Burghardt, and
  Mirmehdi]{masullo2021no}
Alessandro Masullo, Toby~J Perrett, Dima Damen, Tilo Burghardt, and Majid
  Mirmehdi.
\newblock No need for a lab: Towards multi-sensory fusion for ambient assisted
  living in real-world living homes.
\newblock In \emph{16th International Conference on Computer Vision Theory and
  Applications}, 2021.

\bibitem[Masullo et~al.(2022)Masullo, Perrett, Burghardt, Craddock, Damen, and
  Mirmehdi]{masullo2022inertial}
Alessandro Masullo, Toby Perrett, Tilo Burghardt, Ian Craddock, Dima Damen, and
  Majid Mirmehdi.
\newblock Inertial hallucinations--when wearable inertial devices start seeing
  things.
\newblock \emph{arXiv preprint arXiv:2207.06789}, 2022.

\bibitem[Moltisanti et~al.(2019)Moltisanti, Fidler, and
  Damen]{moltisanti2019action}
Davide Moltisanti, Sanja Fidler, and Dima Damen.
\newblock Action recognition from single timestamp supervision in untrimmed
  videos.
\newblock In \emph{Proceedings of the IEEE/CVF Conference on Computer Vision
  and Pattern Recognition}, pages 9915--9924, 2019.

\bibitem[Munro and Damen(2020)]{munro2020multi}
Jonathan Munro and Dima Damen.
\newblock Multi-modal domain adaptation for fine-grained action recognition.
\newblock In \emph{Proceedings of the IEEE/CVF conference on computer vision
  and pattern recognition}, pages 122--132, 2020.

\bibitem[Nag et~al.(2022)Nag, Zhu, Song, and Xiang]{nag2022temporal}
Sauradip Nag, Xiatian Zhu, Yi-Zhe Song, and Tao Xiang.
\newblock Temporal action detection with global segmentation mask learning.
\newblock \emph{arXiv preprint arXiv:2207.06580}, 2022.

\bibitem[Oord et~al.(2018)Oord, Li, and Vinyals]{oord2018representation}
Aaron van~den Oord, Yazhe Li, and Oriol Vinyals.
\newblock Representation learning with contrastive predictive coding.
\newblock \emph{arXiv preprint arXiv:1807.03748}, 2018.

\bibitem[Qing et~al.(2021)Qing, Su, Gan, Wang, Wu, Wang, Qiao, Yan, Gao, and
  Sang]{qing2021temporal}
Zhiwu Qing, Haisheng Su, Weihao Gan, Dongliang Wang, Wei Wu, Xiang Wang,
  Yu~Qiao, Junjie Yan, Changxin Gao, and Nong Sang.
\newblock Temporal context aggregation network for temporal action proposal
  refinement.
\newblock In \emph{Proceedings of the IEEE/CVF conference on computer vision
  and pattern recognition}, pages 485--494, 2021.

\bibitem[Qu et~al.(2021)Qu, Chen, Li, Zhang, Lu, and Knoll]{qu2021acm}
Sanqing Qu, Guang Chen, Zhijun Li, Lijun Zhang, Fan Lu, and Alois Knoll.
\newblock Acm-net: Action context modeling network for weakly-supervised
  temporal action localization.
\newblock \emph{arXiv preprint arXiv:2104.02967}, 2021.

\bibitem[Radford et~al.(2021)Radford, Kim, Hallacy, Ramesh, Goh, Agarwal,
  Sastry, Askell, Mishkin, Clark, et~al.]{radford2021learning}
Alec Radford, Jong~Wook Kim, Chris Hallacy, Aditya Ramesh, Gabriel Goh,
  Sandhini Agarwal, Girish Sastry, Amanda Askell, Pamela Mishkin, Jack Clark,
  et~al.
\newblock Learning transferable visual models from natural language
  supervision.
\newblock In \emph{International Conference on Machine Learning}, pages
  8748--8763. PMLR, 2021.

\bibitem[Shou et~al.(2018)Shou, Gao, Zhang, Miyazawa, and
  Chang]{shou2018autoloc}
Zheng Shou, Hang Gao, Lei Zhang, Kazuyuki Miyazawa, and Shih-Fu Chang.
\newblock Autoloc: Weakly-supervised temporal action localization in untrimmed
  videos.
\newblock In \emph{Proceedings of the European Conference on Computer Vision
  (ECCV)}, pages 154--171, 2018.

\bibitem[Wang et~al.(2022)Wang, Damen, Mirmehdi, and Perrett]{wang2022tvnet}
Hanyuan Wang, Dima Damen, Majid Mirmehdi, and Toby~J Perrett.
\newblock Tvnet: Temporal voting network for action localization.
\newblock In \emph{17th International Conference on Computer Vision Theory and
  Applications}, 2022.

\bibitem[Wang et~al.(2021{\natexlab{a}})Wang, Xing, and
  Liu]{wang2021actionclip}
Mengmeng Wang, Jiazheng Xing, and Yong Liu.
\newblock Actionclip: A new paradigm for video action recognition.
\newblock \emph{arXiv preprint arXiv:2109.08472}, 2021{\natexlab{a}}.

\bibitem[Wang et~al.(2021{\natexlab{b}})Wang, Qing, Huang, Feng, Zhang, Jiang,
  Tang, Gao, and Sang]{wang2021proposal}
Xiang Wang, Zhiwu Qing, Ziyuan Huang, Yutong Feng, Shiwei Zhang, Jianwen Jiang,
  Mingqian Tang, Changxin Gao, and Nong Sang.
\newblock Proposal relation network for temporal action detection.
\newblock \emph{Workshop of the IEEE/CVF conference on computer vision and
  pattern recognition}, 2021{\natexlab{b}}.

\bibitem[Wedel et~al.(2009)Wedel, Pock, Zach, Bischof, and
  Cremers]{wedel2009improved}
Andreas Wedel, Thomas Pock, Christopher Zach, Horst Bischof, and Daniel
  Cremers.
\newblock An improved algorithm for tv-l 1 optical flow.
\newblock In \emph{Statistical and geometrical approaches to visual motion
  analysis}, pages 23--45. Springer, 2009.

\bibitem[Xu et~al.(2021{\natexlab{a}})Xu, Ghosh, Huang, Arora, Aminzadeh,
  Feichtenhofer, Metze, and Zettlemoyer]{xu2021vlm}
Hu~Xu, Gargi Ghosh, Po-Yao Huang, Prahal Arora, Masoumeh Aminzadeh, Christoph
  Feichtenhofer, Florian Metze, and Luke Zettlemoyer.
\newblock Vlm: Task-agnostic video-language model pre-training for video
  understanding.
\newblock \emph{arXiv preprint arXiv:2105.09996}, 2021{\natexlab{a}}.

\bibitem[Xu et~al.(2021{\natexlab{b}})Xu, Ghosh, Huang, Okhonko, Aghajanyan,
  Metze, Zettlemoyer, and Feichtenhofer]{xu2021videoclip}
Hu~Xu, Gargi Ghosh, Po-Yao Huang, Dmytro Okhonko, Armen Aghajanyan, Florian
  Metze, Luke Zettlemoyer, and Christoph Feichtenhofer.
\newblock Videoclip: Contrastive pre-training for zero-shot video-text
  understanding.
\newblock \emph{Proceedings of the 2021 Conference on Empirical Methods in
  Natural Language Processing}, 2021{\natexlab{b}}.

\bibitem[Xu et~al.(2020)Xu, Zhao, Rojas, Thabet, and Ghanem]{xu2020g}
Mengmeng Xu, Chen Zhao, David~S Rojas, Ali Thabet, and Bernard Ghanem.
\newblock G-tad: Sub-graph localization for temporal action detection.
\newblock In \emph{Proceedings of the IEEE/CVF Conference on Computer Vision
  and Pattern Recognition}, pages 10156--10165, 2020.

\bibitem[Yang et~al.(2022{\natexlab{a}})Yang, Wu, Wang, Jin, Xia, Yao, and
  Huang]{yang2022temporal}
Haosen Yang, Wenhao Wu, Lining Wang, Sheng Jin, Boyang Xia, Hongxun Yao, and
  Hujie Huang.
\newblock Temporal action proposal generation with background constraint.
\newblock In \emph{Proceedings of the AAAI Conference on Artificial
  Intelligence}, volume~36, pages 3054--3062, 2022{\natexlab{a}}.

\bibitem[Yang et~al.(2021)Yang, Bisk, and Gao]{yang2021taco}
Jianwei Yang, Yonatan Bisk, and Jianfeng Gao.
\newblock Taco: Token-aware cascade contrastive learning for video-text
  alignment.
\newblock In \emph{Proceedings of the IEEE/CVF International Conference on
  Computer Vision}, pages 11562--11572, 2021.

\bibitem[Yang et~al.(2022{\natexlab{b}})Yang, Huang, Sugano, and
  Sato]{yang2022interact}
Lijin Yang, Yifei Huang, Yusuke Sugano, and Yoichi Sato.
\newblock Interact before align: Leveraging cross-modal knowledge for domain
  adaptive action recognition.
\newblock In \emph{Proceedings of the IEEE/CVF Conference on Computer Vision
  and Pattern Recognition}, pages 14722--14732, 2022{\natexlab{b}}.

\bibitem[Yang et~al.(2023)Yang, Kong, Yang, Kehl, Sato, and
  Kobori]{yang2023deco}
Lijin Yang, Quan Kong, Hsuan-Kung Yang, Wadim Kehl, Yoichi Sato, and Norimasa
  Kobori.
\newblock Deco: Decomposition and reconstruction for compositional temporal
  grounding via coarse-to-fine contrastive ranking.
\newblock In \emph{Proceedings of the IEEE/CVF Conference on Computer Vision
  and Pattern Recognition}, pages 23130--23140, 2023.

\bibitem[Yu et~al.(2023)Yu, Huang, Furuta, Yagi, Goutsu, and Sato]{yu2023fine}
Zecheng Yu, Yifei Huang, Ryosuke Furuta, Takuma Yagi, Yusuke Goutsu, and Yoichi
  Sato.
\newblock Fine-grained affordance annotation for egocentric hand-object
  interaction videos.
\newblock In \emph{Proceedings of the IEEE/CVF Winter Conference on
  Applications of Computer Vision}, pages 2155--2163, 2023.

\bibitem[Zeng et~al.(2021)Zeng, Huang, Tan, Rong, Zhao, Huang, and
  Gan]{zeng2021graph}
Runhao Zeng, Wenbing Huang, Mingkui Tan, Yu~Rong, Peilin Zhao, Junzhou Huang,
  and Chuang Gan.
\newblock Graph convolutional module for temporal action localization in
  videos.
\newblock \emph{IEEE Transactions on Pattern Analysis and Machine
  Intelligence}, 44\penalty0 (10):\penalty0 6209--6223, 2021.

\bibitem[Zhang et~al.(2021)Zhang, Cao, Yang, Chen, and Zou]{zhang2021cola}
Can Zhang, Meng Cao, Dongming Yang, Jie Chen, and Yuexian Zou.
\newblock Cola: Weakly-supervised temporal action localization with snippet
  contrastive learning.
\newblock In \emph{Proceedings of the IEEE/CVF Conference on Computer Vision
  and Pattern Recognition}, pages 16010--16019, 2021.

\bibitem[Zhao et~al.(2021)Zhao, Thabet, and Ghanem]{zhao2021video}
Chen Zhao, Ali~K Thabet, and Bernard Ghanem.
\newblock Video self-stitching graph network for temporal action localization.
\newblock In \emph{Proceedings of the IEEE/CVF International Conference on
  Computer Vision}, pages 13658--13667, 2021.

\end{thebibliography}
\end{document}